\documentclass[letterpaper, 10 pt, conference]{ieeeconf}  
\IEEEoverridecommandlockouts                              
\usepackage{cite}
\usepackage{amsmath}
\usepackage{amssymb}
\usepackage{dsfont}
\usepackage{graphicx}
\usepackage{subcaption}
\usepackage[nolist]{acronym}
\usepackage{cleveref}
\usepackage{amssymb}
\usepackage{bm}
\crefname{figure}{Fig.}{Figs.}
\usepackage[subrefformat=parens,labelformat=parens]{subcaption}
\usepackage{url}
\usepackage{xcolor}
\usepackage[utf8]{inputenc}
\usepackage[utf8]{inputenx}
\usepackage[T1]{fontenc}
\DeclareUnicodeCharacter{0160}{\v{S}}  % Š
\DeclareUnicodeCharacter{0161}{\v{s}}  % š
\usepackage{multirow}
\usepackage{url}
\usepackage{tikz}
\usepackage{cleveref}
\crefname{figure}{Fig.}{Figs.}
\crefname{section}{Sec.}{Secs.}
\crefname{equation}{Eq.}{Eqs.}
\crefname{table}{Table}{Table}
\usepackage{tabularx}
\usepackage{csquotes}
\usepackage{tcolorbox}
\usepackage{multirow}
\usepackage{booktabs}
\tcbset{
  colback=yellow!20, 
  colframe=black,    
  boxrule=0.5pt,
  left=1mm,
  right=1mm,
  top=1mm,
  bottom=1mm
}
\newcommand{\RGB}{\mathbf{c}}
\newcommand{\RGBcurrent}{\mathbf{c}_{\mathrm{current}}}
\newcommand{\RGBtarget}{\mathbf{c}_{\mathrm{target}}}
\newcommand{\RGBadded}{\mathbf{c}_{\mathrm{added}}}

\begin{acronym}
\acro{RL}[RL]{Reinforcement Learning}
\acro{DRL}[DRL]{Deep Reinforcement Learning}
\acro{mc}[MC]{Mountain Car}
\acro{iqm}[IQM]{Interquartile Mean}
\acro{nn}[NN]{Neural Network}
\acro{dqn}[DQN]{Deep Q-Network}
\acro{mdp}[MDP]{Markov Decision Process}
\acro{lerp}[Lerp]{Linear Interpolation}
\acro{ppo}[PPO]{Proximal Policy Optimization}
\acro{KM}[KM]{Kubelka-Munk}
\acro{WGM}[WGM]{Weighted Geometric Mean}
\acro{icps}[ICPS]{Industrial Cyber-Physical System}
\acro{ARL}[ARL]{Adversarial Reinforcement Learning}
\acro{FP}[FP]{Final Performance}
\acro{T7.5}[T7.5]{Time-to-7.5 Performance}
\acro{CV}[CV]{Coefficient of Variation}
\acro{NM}[NM]{Non-Monotonicity}
\acro{CS}[CS]{Composite Score}
\end{acronym}

\title{\LARGE \bf
Impact of Markov Decision Process Design on Sim-to-Real Reinforcement Learning
}

\author{Tatjana Krau, Jorge Mandlmaier, Tobias Damm, Frieder Heieck%
\thanks{This work was supported by the Federal Ministry of Education and Research Germany under the project iCARus (13GW0657B).}
\thanks{T. Krau, J. Mandlmaier, F. Heieck are with Institute of Production and Informatics,
        87527 Sonthofen, Germany
        {\tt\small [tatjana.krau,jorge.mandlmaier,\linebreak
        frieder.heieck]@hs-kempten.de}}%
\thanks{T. Damm is with the Department of Mathematics, RPTU University Kaiserslautern-Landau,
        67663 Kaiserslautern, Germany
        {\tt\small t.damm@rptu.de}}%
}
\begin{document}

\maketitle
\thispagestyle{empty}
\pagestyle{empty}

%%%%%%%%%%%%%%%%%%%%%%%%%%%%%%%%%%%%%%%%%%%%%%%%%%%%%%%%%%%%%%%%%%%%%%%%%%%%%%%%
\begin{abstract}
\ac{RL} has demonstrated strong potential for industrial process control, yet policies trained in simulation often suffer from a significant sim-to-real gap when deployed on physical hardware. This work systematically analyzes how core \ac{mdp} design choices—state composition, target inclusion, reward formulation, termination criteria, and environment dynamics models—affect this transfer. Using a color mixing task, we evaluate different \ac{mdp} configurations and mixing dynamics across simulation and real-world experiments.  We validate our findings on physical hardware, demonstrating that physics-based dynamics models achieve up to 56\% real-world success under strict precision constraints where simplified models fail entirely. Our results provide practical \ac{mdp} design guidelines for deploying \ac{RL} in industrial process control.
\end{abstract}
\acresetall
%%%%%%%%%%%%%%%%%%%%%%%%%%%%%%%%%%%%%%%%%%%%%%%%%%%%%%%%%%%%%%%%%%%%%%%%%%%%%%%%
\section{INTRODUCTION}\label{sec:intro}
\ac{RL} provides a structured approach for sequential decision-making in settings where explicit system models are unavailable or difficult to construct~\cite{sutton_reinforcement_2018}. Model-free \ac{RL} methods derive policies directly from interaction data, making them suitable for nonlinear and stochastic systems with complex dynamics~\cite{sutton_reinforcement_2018}. Recent developments in deep \ac{RL} extend these techniques to high-dimensional continuous state and action spaces, enabling applications in robotics, energy management, and industrial process control.

Despite these advances, policies trained in simulation frequently exhibit significant performance degradation when deployed on physical systems. This \emph{sim-to-real gap} arises from discrepancies between simulated and real-world environments, causing policies to yield suboptimal or unsafe behavior~\cite{hofer_sim2real_2021}. For industrial process control, these factors are particularly consequential: safety-critical constraints preclude extensive real-world exploration, and stringent precision requirements amplify the impact of even small policy errors.

To investigate strategies for bridging this gap, we adopt \textbf{color mixing} as a \emph{physical testbed} that enables controlled, reproducible experiments on real hardware. Unlike simulation-only studies, we validate the findings through hardware deployment, directly measuring the theory-to-practice gap that limits industrial RL adoption. Real-world variations in lighting, pigment properties, and dispensing accuracy create a substantial sim-to-real gap even for this apparently simple task. A target application motivating this work is CAR-T cell therapy, where \ac{RL} could automate aspects of the costly, patient-specific treatment process—goals pursued by the research project iCARus~\footnote{https://kefis.fza.hs-kempten.de/de/forschungsprojekt/507-icarus}. The analogy is instructive: just as base colors are combined in precise proportions to achieve a target color, CAR-T production involves the controlled mixing of fluids with specific concentrations to reach optimal cellular conditions. Achieving a target color in the testbed thus mirrors the precision required in cell culture preparation. This research focuses exclusively on \ac{RL} methodology development and does not involve biological systems.

Predominant approaches to reducing the sim-to-real gap focus on \emph{transition dynamics}: domain randomization trains policies across simulator parameter distributions~\cite{hofer_sim2real_2021}, while system identification methods align dynamics models. However, these approaches hold other \ac{mdp} components fixed. Yet an \ac{mdp} comprises not only transition probabilities but also states, rewards, and termination criteria~\cite{sutton_reinforcement_2018}. Prior work on reward shaping, curriculum learning, and state abstraction has examined these components for learning efficiency, but with limited analysis of their effect on \emph{sim-to-real transfer}.
Recently, Schäfer et al.~\cite{schafer_crucial_2025} highlighted the crucial role of systematic \ac{mdp} formulation for real-world RL and validated their insights on physical hardware. Inspired by this concept, we adapt the idea of structured \ac{mdp} analysis to our color-mixing testbed.\\
\textbf{Scope and Positioning.} This work is primarily empirical and application-driven. While theoretical analyses of sim-to-real transfer require restrictive assumptions on dynamics mismatch~\cite{recht_tour_2019,gros_datadriven_2020}, our contribution lies in extracting actionable design principles from systematic hardware experiments. The color mixing domain provides a controlled setting where the sim-to-real gap is measurable and reproducible.

This enables a systematic study of how different \ac{mdp} design choices affect transfer performance. Accordingly, we ask: \textit{How do \ac{mdp} design choices influence the transferability of \ac{mdp} policies from simulation to physical systems?} We address this through a structured empirical study with the following contributions:
\begin{enumerate}
    \item We systematically study how \ac{mdp} design choices—state representation, reward shaping, termination criteria, and dynamics fidelity, affect training stability and sim-to-real transfer in a color-mixing testbed.
    \item We empirically quantify the sim-to-real gap across these configurations, revealing which design factors most strongly impact transfer performance.
    \item We identify \ac{mdp} design patterns that improve policy transferability and expose failure modes caused by formulation-induced overfitting, providing actionable guidance for real-world RL.
\end{enumerate}

\section{BACKGROUND}\label{sec:background}
We formalize the color mixing task as a finite-horizon \ac{mdp} defined by the tuple $(\mathcal{S}, \mathcal{A}, P, R, T)$, where $\mathcal{S} \subseteq \mathbb{R}^d$ denotes the state space, $\mathcal{A} = \{1, \ldots, |\mathcal{A}|\}$ a \emph{finite} action space, $P: \mathcal{S} \times \mathcal{A} \times \mathcal{S} \to [0,1]$ the transition probability function, $R: \mathcal{S} \times \mathcal{A} \to \mathbb{R}$ the reward function, and $T \in \mathbb{N}$ the \textbf{termination horizon}. The specific formulations of $\mathcal{S}$, $\mathcal{A}$, and $R$ constitute the primary experimental variables of this study and are detailed in ~\cref{sec:systematic_mdp}.

\subsection{Color Mixing as a Reinforcement Learning Task}
\begin{figure}[!ht]
    \centering
    \includegraphics[width=0.3\textwidth]{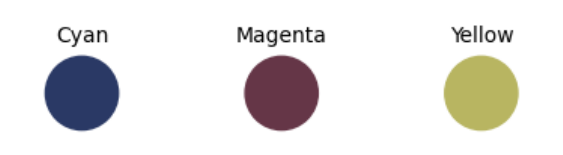}
    \caption{Visual representation of the three base ink colors: cyan, magenta and yellow.}
    \label{fig:cmy}
\end{figure}

We formulate color mixing as an episodic \ac{RL} task following the setup introduced in~\cite{krau_exploring_variance}. All colors are represented in the RGB color space as vectors $\RGB \in [0,255]^3$. The agent incrementally mixes three physical printer inks, commercially labeled as cyan, magenta, and yellow, with the objective of reproducing a predefined target color. When measured and expressed in the RGB color space, these inks correspond to fixed empirical color vectors, namely cyan $[42,57,101]$, magenta $[101,54,71]$, and yellow $[184,181,97]$ (see~\cref{fig:cmy}). Importantly, these values do not represent idealized RGB primaries but capture the non-ideal, real-world pigment characteristics of the inks.

The similarity between the current mixture  $\RGBcurrent$ and the target color $\RGBtarget$ is quantified using the Euclidean distance in RGB space:
\begin{equation}\label{equ:EucliDistance}
    d = \|\RGBcurrent - \RGBtarget\|_2
\end{equation}
We use Euclidean distance in RGB space rather than perceptually uniform metrics (e.g., CIELAB $\Delta E$) for two reasons: (i)~RGB values are directly measured by our camera without colorimetric transformations that introduce calibration errors, and (ii)~RGB distance is consistent between simulation and hardware, enabling direct gap quantification.

The target color is considered reached if the absolute deviation in each RGB channel is below the \textbf{tolerance~$\tau$}:
\begin{equation}\label{equ:TargetReached}
    |(\RGBcurrent)_k - (\RGBtarget)_k| \leq \tau, \quad \forall k \in \{R,G,B\}.
\end{equation}

\subsection{Color Prediction Models}\label{Sec:ColorPredModel}
Predicting the resulting color from mixing multiple pigments is non-trivial, as color blending is inherently non-linear due to light absorption and scattering effects. To study the influence of color modeling fidelity on \ac{RL} performance, we compare three color prediction models with increasing physical realism:

\begin{itemize}
    \item \textbf{\ac{lerp}}: A simple linear interpolation model serving as a computationally cheap but physically unrealistic baseline~\cite{sochorova_practical_2021}.
    \item \textbf{\ac{KM}}: A physics-based model derived from the \ac{KM} theory, capturing absorption and scattering in diffusely reflective materials~\cite{sochorova_practical_2021}. We implement this model using the Mixbox library\footnote{\url{https://github.com/scrtwpns/mixbox}}, which provides a practical approximation suitable for real-time simulation.
   \item \textbf{\ac{WGM}}: A purely subtractive spectral mixing model based on the weighted geometric mean of reflectance spectra~\cite{simonet_WGM}. For $n$ colors with mixing volumes $V_i$ and spectral representations $S_i(\lambda)$, $\lambda \in [380, 780]$\,nm, the mixed spectrum at wavelength $\lambda$ is given by:
    \begin{equation}
        S_{\mathrm{mixed}}(\lambda) = \prod_{i=1}^{n} S_i(\lambda)^{w_i}, \quad \text{where } w_i = \frac{V_i}{\sum_{j=1}^{n} V_j}.
    \end{equation}
\end{itemize}

\subsection{Robustness Mechanisms}\label{Sec:Robustness}
To support stable sim-to-real transfer, we include two fixed robustness mechanisms, kept constant across all \ac{mdp} formulations. 
First, we add small channel-wise perturbations to the RGB observations. The per-channel standard deviations are estimated from repeated real-world measurements of static color targets and kept fixed during training. This exposes the agent to realistic measurement noise, including slight variation across steps. 
Second, we apply a lightweight adversarial perturbation scheme inspired by \ac{ARL}~\cite{huang_adversarial_2017}. Instead of training a separate adversarial policy, we inject bounded worst-case perturbations into the RGB observation with probability 
$p_{\mathrm{adv}} = 0.8$ per step.

\subsection{Metrics}\label{sec:Metrics}
We employ separate metrics for simulation training and real-world evaluation to capture both learning efficiency and transfer quality.
\subsubsection{Simulation Metrics}
Each configuration is evaluated using four metrics derived from the episode reward mean curve over training:
\begin{itemize}
    \item \textbf{\ac{FP}:} Mean episode reward over the last 10\% of training steps, capturing asymptotic policy quality.
    \item \textbf{\ac{T7.5}:} Number of training steps required to reach a mean reward of 7.5, measuring sample efficiency. %The threshold 7.5 corresponds to the typical performance of a baseline agent after moderate training, used to measure sample efficiency.
    \item \textbf{\ac{CV}:} Ratio of standard deviation to mean reward over the last 20\% of training, quantifying late-stage stability.
    \item \textbf{\ac{NM}:} Number of reward drops exceeding 5\% between consecutive 5000-step windows, measuring susceptibility to catastrophic forgetting.
\end{itemize}
To rank configurations, we computed a weighted \ac{CS}. \ac{FP} (40\%) and \ac{CV} (30\%) were prioritized due to their relevance for sim-to-real transfer, followed by \ac{T7.5} (20\%) and \ac{NM} (10\%).

\subsubsection{Real-world Metrics} 
Hardware experiments are evaluated using:
\begin{itemize}
    \item \textbf{RGB Distance} ($d_{\mathrm{R}}$): Euclidean distance between the achieved and target RGB values per~\cref{equ:EucliDistance}, capturing final color accuracy.
    \item \textbf{Steps-to-Target} ($s_{\mathrm{R}}$): Number of interaction steps required to reach the target color, measuring real-world efficiency.
    \item \textbf{Success}: Proportion of episodes achieving the tolerance criterion.
\end{itemize}

\section{SYSTEMATIC \ac{mdp} OPTIMIZATION}\label{sec:systematic_mdp}
We adopt a phase-wise optimization strategy to systematically improve \ac{mdp} design. Each phase isolates specific design choices—reward functions, state composition, episode parameters, and dynamics models allowing us to attribute performance improvements to individual components.

\subsection{Phase 1: Component Selection}\label{Phase1}
\textit{Objective:} Evaluate how state composition, target state inclusion, and reward design affect learning stability and generalization.

\textit{Fixed Parameters:} 
To minimize computational cost and focus on structural effects, we fix:
\begin{itemize}
    \item Dynamics model: \ac{lerp}
    \item Termination horizon $T$: 20 steps
    \item Tolerance $\tau$: 10,~\cref{equ:TargetReached}
    \item Observation robustness: channel-wise noise and adversarial perturbations~(see~\cref{Sec:Robustness})
    \item Total training time: 500k steps
\end{itemize}

\subsection*{Target State Inclusion}
Since each episode samples a different target color $\RGBtarget$, we investigate the effect of including it in the
observations. This design choice has theoretical implications rooted in the Markov property and goal-conditioned
\ac{RL}~\cite{sutton_reinforcement_2018,schaul_universal_2015}.

\textbf{Theoretical Motivation.}
Consider a reward function $r(s_t, a_t, g)$ that evaluates performance relative to a goal $g \in \mathcal{G}$. When
$g$ is \emph{not} included in the agent's state, the policy $\pi(a|s)$ cannot condition on the goal. The agent must
therefore optimize the expected return marginalized over the goal distribution:
\begin{equation}\label{eq:marginalized_objective}
J(\pi) = \mathbb{E}_{g \sim p(g)} \left[ \mathbb{E}_{\tau \sim \pi} \left[ \sum_{t=0}^{T} \gamma^t r(s_t, a_t, g)
\right] \right].
\end{equation}
By linearity of expectation, this is equivalent to optimizing the average reward function $\bar{r}(s,a) =
\mathbb{E}_{g}[r(s,a,g)]$. The resulting policy converges to a single compromise behavior that performs moderately
across all goals but cannot achieve any specific goal optimally~\cite{kaelbling_learning_1993}.

In contrast, including $g$ in the state space yields $\tilde{\mathcal{S}} = \mathcal{S} \times \mathcal{G}$, enabling
goal-conditioned policies $\pi(a|s,g)$. The agent can then learn distinct optimal strategies for each
target~\cite{schaul_universal_2015}.

\begin{tcolorbox}[title=H1: Target State Inclusion, colback=yellow!8, colframe=black, fonttitle=\bfseries]
Including the target color in the state is essential for sim-to-real transfer. Omitting $\RGBtarget$ violates the Markov property—the reward depends on information unavailable in the state—transforming the \ac{mdp} into a POMDP~\cite{ghosh_why_2021} and causing the agent to learn a compromise policy that fails under real-world dynamics.
\end{tcolorbox}

We test this hypothesis by comparing two variants: one where $\RGBtarget$ is part of the state, and one where it is omitted.

\subsection*{State Composition}
The volume representation in the state space is critical, as it 
determines how the agent perceives mixing ratios.
\begin{tcolorbox}[title=H2 (State Representation), colback=yellow!8, colframe=black, fonttitle=\bfseries]
State representations encoding relative color proportions generalize better than absolute volume representations, as ratio-based encodings are invariant to scale and total volume.
\end{tcolorbox}
We evaluate five state variants (\cref{tab:state_representations}). Based on the state representation, the \textit{finite action space} consists of selecting one of three colors with either absolute volumes ($20$--$200\,\mu$l in increments of 20) for states representation~0 and~1, or relative fractions ($0.1$--$1.0$) for states representations~2--4.

\begin{table}[t]
\label{tab:states}
    \centering
    \caption{State representations evaluated in this study. All states include $\RGBcurrent \in [0,255]^3$ and the volume encoding. States with target inclusion additionally contain $\RGBtarget \in [0,255]^3$. Example mixture: $200\,\mu$l Cyan + $20\,\mu$l Magenta.}
    \label{tab:state_representations}
    \begin{tabular}{@{} p{0.1cm} p{2.9cm} p{1.7cm} p{2.5cm} @{}}
        \toprule
        ID & Volume Encoding & Example & Description \\
        \midrule
        0 & $V_{\mathrm{sum}}$ & $200\,\mu$l & Absolute total volume \\
        1 & $[V_1, V_2, V_3]$ & $[200, 20, 0]\,\mu$l & Absolute per-channel \\
        2 & $V_{\mathrm{sum,rel}}$ & $1.1$ & Relative total volume \\
        3 & $[V_{1,\mathrm{rel}}, V_{2,\mathrm{rel}}, V_{3,\mathrm{rel}}]$ & $[1.0, 10.1, 0]$ & Relative per-channel \\
        4 & $[V_{1,\mathrm{ratio}}, V_{2,\mathrm{ratio}}, V_{3,\mathrm{ratio}}]$ & $[0.909, 0.091, 0]$ & Normalized ratios \\
        \bottomrule
    \end{tabular}
\end{table}

\subsection*{Reward Functions}
We evaluate three reward formulations. In all cases, the agent receives an additional bonus of $+10$ whenever the target color is reached:
\paragraph{R1 (Normalized Euclidean distance)}
\[
 r_t = -\frac{\|\RGBcurrent - \RGBtarget\|_2}{d_{\max}} +\mathds{1}_{\text{success}} \cdot 10
\]
where $d_{\max} = \sqrt{3} \cdot 255 \approx 441.7$ is the maximum possible Euclidean distance in RGB space. This reward encourages the agent to approach the target color by directly reflecting the distance in the RGB space.
\paragraph{R2–R3 (Action Penalty)}
\[
 r_t = -\frac{\|\RGBadded - \RGBtarget\|_2}{D} \cdot V^2 
 + \mathds{1}_{\text{success}} \cdot 10
\]
Both variants additionally penalize large dispensing volumes and suboptimal color choices.  
The normalization constant $D$ differs:  
for R2, $D = \max_{j,k} d_{jk}$ (maximum pairwise distance between base colors);  
for R3, $D = \max_j d_j$ (maximum distance from any base color to the target).

To compare how the complexity of the reward function affects training stability and sim-to-real transfer, we formulate the following hypothesis:
\begin{tcolorbox}[title=H3 (Reward Complexity), colback=yellow!8, colframe=black, fonttitle=\bfseries]
Simple, distance-based reward functions yield more stable training and superior sim-to-real transfer than action-penalized rewards, which introduce inductive bias that may overfit to simulation-specific dynamics.
\end{tcolorbox}

\subsection{Phase 2: Episode Design Optimization}
\textit{Objective:} Identify optimal termination horizon $T$ and distance tolerance $\tau$ in simulation to balance convergence speed and precision.

\textit{Fixed Parameters:} To isolate the effect of episode design, we fix:  
\begin{itemize}
    \item Best-performing state representation, reward function, and target inclusion from Phase~1  
    \item Dynamics model, observation noise and \ac{ARL} unchanged  
    \item Total training time: 250k steps, since Phase~2 investigates local effects of termination and tolerance parameters and does not require full policy relearning.  
\end{itemize}

\textit{Parameters Tested:}
\begin{itemize}
    \item \textbf{Termination $T$} $\in \{5,10,20,30\}$ 
    \item \textbf{Tolerance $\tau$} $\in \{2.5,\, 5,\, 7.5\}$, compared to~\cref{equ:TargetReached}
\end{itemize}

Based on these tests, we hypothesize the following relationship between training thresholds and performance:
\begin{tcolorbox}[title=H4 (Termination and Tolerance Criteria), colback=yellow!8, colframe=black, fonttitle=\bfseries]
Strict training thresholds reduce success but promote precision, while relaxed deployment thresholds accommodate real-world noise and variability.
\end{tcolorbox}

\subsection{Phase 3: Dynamics Robustness}
\textit{Objective:} Test whether the optimized \ac{mdp} generalizes to more realistic prediction models, specifically \ac{KM} from~\cref{Sec:ColorPredModel}.

\textit{Fixed Parameters:} 
\begin{itemize}
    \item Best-performing configurations from Phase~1 and 2
    \item Total training time: 500k steps
\end{itemize}

\textit{Parameters Tested:}
\begin{itemize}
    \item Dynamics model: \ac{KM}
\end{itemize}

\begin{tcolorbox}[title=H5 (Dynamics Model), colback=yellow!8, colframe=black, fonttitle=\bfseries]
Realistic mixing models increase task complexity but better reflect real-world behavior; success under complex dynamics indicates robustness when transferred to physical experiments.
\end{tcolorbox}

\begin{table}[t]
    \centering
    \caption{Overview of experimental phases. Left column lists parameters, middle column shows tested values, right column indicates selected configuration for subsequent phases.}
    \label{tab:phase1_phase2_phase3}
    \begin{tabular}{@{}lll@{}}
        \toprule
        \textbf{Phase 1} & Tested & Selected \\
        \midrule
        \multicolumn{3}{@{}l@{}}{\textit{Fixed: \ac{lerp}, Noise, \ac{ARL}, $T{=}20$, $\tau{=}10$}} \\
        Target inclusion & yes / no & yes \\
        State & 0 / 1 / 2 / 3 / 4 & 2, 3, 4 \\
        Reward & R1 / R2 / R3 & R1, R2 \\
        \midrule
        \textbf{Phase 2} & Tested & Selected \\
        \midrule
        \multicolumn{3}{@{}p{8.5cm}@{}}{\textit{Fixed: \ac{lerp}, Noise, \ac{ARL}, Target yes, State $\in \{2,3,4\}$, Reward $\in \{\text{R1,R2}\}$}} \\
        Termination $T$ & 5 / 10 / 20 / 30  & 5 \\
        Tolerance $\tau$ & 2.5 / 5 / 7.5 & 7.5 \\
        \midrule
        \textbf{Phase 3} & Tested & Selected \\
        \midrule
        \multicolumn{3}{@{}p{8.5cm}@{}}{\textit{Fixed: Noise, \ac{ARL}, Target yes, State $\in \{2,3,4\}$, Reward $\in \{\text{R1,R2}\}$, $T{=}5$, $\tau{=}7.5$}} \\
        Dynamics & \ac{KM}  & R1, State 4 \\
        \midrule
        \textbf{Final Config.} & \multicolumn{2}{p{5.5cm}}{Noise, \ac{ARL}, State 4, R1, Target=yes, $T{=}5$, $\tau{=}7.5$, WGM} \\
        \bottomrule
    \end{tabular}
\end{table}

\section{EXPERIMENTAL SETUP}\label{sec:experimental setup}
\subsection{Simulation Environment}
\cref{tab:phase1_phase2_phase3} summarizes the progression through all optimization phases and the resulting final configuration. The simulation setup used to evaluate these configurations follows the workflow in~\cite{krau_exploring_variance}: starting from the base color (out of the three available) with the smallest Euclidean distance to the target color and the same amount, the agent incrementally adds one of three base colors, predicts the resulting mixture~(\cref{Sec:ColorPredModel}), and receives rewards~(\cref{Phase1}). In this work, we additionally inject camera noise and \ac{ARL} to evaluate robustness~(\cref{Sec:Robustness}). Training uses \ac{ppo}~\cite{schulman_proximal_2017}, a policy gradient method that constrains policy updates for stable learning, implemented in Stable-Baselines3 with default hyperparameters~\cite{raffin_stable-baselines3_2021}.

\subsection{Hardware Evaluation Protocol}\label{sec:HWEvaluationProtocol}
For real-world evaluation, the agent executes actions on a controlled setup: the selected base color is added to the current mixture on a flat surface using a standardized pipetting and stirring procedure. Images are captured under a webcam in a light-controlled photo box to ensure consistent illumination, and RGB values are measured after a fixed time interval. Each episode starts from the base color closest to the target and proceeds for a maximum of five mixing steps with a tolerance of $\tau=7.5$. 

Each policy is tested on 4 target colors with 4 repetitions. Targets cover diverse chromatic regions: C1 [128,91,67], C2 [42,76,66], C3 [39,52,56], and C4 [67,64,75], pre-mixed to ensure physical achievability.

Only the configuration listed in~\cref{Tab:Testmodels_HW} is evaluated on hardware.
\begin{table}[t]
\centering
\caption{Models evaluated on hardware. Phase indicates where the configuration was determined. The remaining columns show the training parameters.}
\label{Tab:Testmodels_HW}
\resizebox{\columnwidth}{!}{
\begin{tabular}{l|c|c|c|c|c|c|c}
\toprule
\textbf{Model} &\textbf{Phase} & \textbf{Target}& \textbf{State} & \textbf{Termination $T$} & \textbf{Tolerance $\tau$} & \textbf{Reward} & \textbf{Dynamics Model} \\
\midrule
M1 & 1   & Yes & 4 & 20 & 10 & R1 & \ac{lerp}\\
M2 & 1  & No & 4 & 20 & 10 & R1 & \ac{lerp}\\
M3 & 2   & Yes & 4 & 5 & 7.5 & R1 & \ac{lerp}\\
M4 & 3   & Yes & 4 & 5 & 7.5 & R1 & \ac{KM}\\
M5 & 3  & Yes & 4 & 5 & 7.5 & R1 & \ac{WGM}\\
\bottomrule
\end{tabular}}
\end{table}

\section{RESULTS}
\subsection{Simulation}
Based on the metrics defined in \cref{sec:Metrics}, we evaluated all configurations from~\cref{tab:phase1_phase2_phase3}.~\cref{tab:phasenwerte} reports the performance metrics for key configurations discussed below.
\subsubsection{H1 (Target State Inclusion)}
 M1 (target included) achieves higher \ac{FP} and substantially lower \ac{CV} than M2 (target excluded). 

\textbf{H1 Verdict:} \emph{Supported.} Target inclusion improves learning stability and convergence.

\subsubsection{H2 (State Representation) \& H3 (Reward Complexity)}
Under KM dynamics, State~4 with a simple Euclidean reward (R1) led to stable learning, whereas other state representations or action-penalized rewards often resulted in unstable policies or near-complete failure.  

\textbf{H2 Verdict:} \emph{Supported.} Normalized ratio-based representations (State~4) generalize best.

\textbf{H3 Verdict:} \emph{Supported.} Simple distance-based rewards outperform action-penalized alternatives under complex dynamics.

\subsubsection{H4 (Termination and Tolerance Criteria)}
Stricter thresholds ($T=5$, $\tau=7.5$) reduce training success compared to relaxed settings ($T=10$, $\tau=7.5$), with \ac{FP} dropping from $9.14 \pm 0.06$ to $7.30 \pm 0.14$. However, we adopt the stricter configuration to prepare the agent for precision requirements in deployment.

\textbf{H4 Verdict:} \emph{Partially supported.} Stricter criteria reduce simulation success.

\subsubsection{H5 (Dynamics Model)} 
\cref{tab:phasenwerte} shows that \ac{lerp} converges fastest (\ac{T7.5} $\approx$ 15k), while \ac{KM} is slowest (\ac{T7.5} $>$ 200k) with higher instability. \ac{WGM} offers a middle ground (\ac{T7.5} $\approx$ 56k).

\textbf{H5 Verdict:} \emph{Supported.} Complex dynamics slow learning but may improve transfer (evaluated in~\cref{sec:real_results}).

\begin{table}[h] % Positionierung hier (h=here)
\centering
\scriptsize
\caption{Simulation performance metrics for key configurations based on episode reward mean. $\uparrow$: higher is better; $\downarrow$: lower is better.}
\begin{tabular}{l|l|l|l|l|l}
\toprule
\textbf{Phase} & \textbf{\ac{FP}}$\uparrow$ & \textbf{\ac{CV}}$\downarrow$ & \textbf{\ac{T7.5}}$\downarrow$ & \textbf{\ac{NM}}$\downarrow$ & \textbf{\ac{CS}}$\uparrow$ \\
\midrule
1 - Target yes & 9.75 $\pm$ 0.02 & 0.0119 & 15,018 & 0 & 0.998 \\%24

1 - Target no & 6.99 $\pm$ 0.23 & 0.07 & 178,858 & 20 & 0.93\\%4
\cmidrule(lr){1-6}
2 - 10, 7.5 & 9.14 $\pm$ 0.06 & 0.03 & 28,672 & 1.7 & 0.94  \\%38

2 - 5, 7.5 & 7.3 $\pm$ 0.14 & 0.06 & 92,842 & 14 & 0.75 \\%71
\cmidrule(lr){1-6}
3 - R1, State 4 & 6.81 $\pm$ 0.02 & 0.06 & 209,578 & 20.7 & 0.77 \\%0

3 - R2 & 6.01 $\pm$ 0.05 & 0.08 & -- & 21.3 & 0.71 \\%1

3 - State 2 & 0.04 $\pm$ 0.02 & 0.19 & -- & 34 & 0.19 \\%4

3 - State 3 & -0.18 $\pm$ 0.05 & 0.98 & -- & 4.3 & 0.54 \\%5
\cmidrule(lr){1-6}
Final Parameters & 7.38 $\pm$ 0.01 & 0.05 & 55,978 & 17 & 0.5\\
\bottomrule
\end{tabular}
\label{tab:phasenwerte}
\end{table}

\subsection{Reality}\label{sec:real_results}
~\cref{tab:results_detailed} summarizes hardware performance across all models from~\cref{Tab:Testmodels_HW} and target colors. Due to space limitations, the table reports only the representative constraint settings C1 and C4, reflecting medium and highly restrictive control regimes, respectively. The intermediate settings C2 and C3 are omitted for brevity, as C2 was consistently solved across models and C3 exhibited failure behavior comparable to C4. The \textbf{Avg} column reflects the aggregated performance across all constraint settings (C1–C4).

\subsubsection{H1: Target State Inclusion}
M2 fails entirely despite moderate simulation performance, while M1 achieves 43.75\% success. This dramatic gap confirms the theoretical prediction from~\cref{eq:marginalized_objective}: without access to $\RGBtarget$, M2 learned a policy optimizing for the average target color. In simulation, where dynamics
match exactly, this compromise policy still accumulates reasonable rewards. However, when real-world dynamics shift
the achievable color space, the agent lacks the goal information needed to adapt its strategy, causing complete
failure.

\textbf{H1 Verdict:} \emph{Strongly supported.} Moderate simulation performance does not guarantee transfer when goal
information is not provided. This finding aligns with goal-conditioned RL
theory~\cite{schaul_universal_2015,kaelbling_learning_1993}.

\subsubsection{H4: Termination and Tolerance}
H4 hypothesizes that stricter episode parameters promote precision but may reduce success. This is observed for low-fidelity dynamics (M3), which fail under strict settings, but enable moderate success with the same dynamics and relaxed parameters. This indicates a clear relationship between termination and tolerance values and the underlying dynamics: in contrast to \ac{lerp}, higher-fidelity models (M4 and M5) perform well under strict parameters. 

\textbf{H4 Verdict:} \emph{Partially supported; effects depend on dynamics model.} The key insight is that episode design interacts with dynamics fidelity: strict thresholds are beneficial only if the model can support precise control.

\subsubsection{H5: Dynamics Model Fidelity}
H5 states that physics-based dynamics improve sim-to-real transfer. Under identical strict parameters, M4 (KM) and M5 (WGM) each succeed in 9 of 16 trials (56.25\%) and reach the lowest final distances of all five models, whereas the Lerp-based M3 fails on every target. M1 (Lerp with relaxed parameters) reaches 43.75\%, showing that relaxed episode constraints can partially offset simple dynamics.

\textbf{H5 Verdict:} \emph{Supported.} Physically grounded models enable robust transfer under precision constraints.

\subsubsection{Color Spectrum Reachability}\label{sec:reachability}
A systematic analysis of the simulation models reveals a fundamental limitation: the four target colors used in hardware experiments lie \emph{outside the producible range} of all three dynamics models in simulation. For each model, we performed an exhaustive search over all possible mixing weight combinations based on the action space to find the closest achievable color to each target. \cref{tab:reachability} shows the minimum per-channel tolerance $\tau_{\min}$ required for each target to be considered ``reached'' under each dynamics model.

\begin{table}[h]
\centering
\caption{Minimum required tolerance $\tau_{\min}$ for each target color to be reachable under each dynamics model.}
\label{tab:reachability}
\begin{tabular}{l ccc}
\toprule
\textbf{Target} & \textbf{\ac{lerp}} & \textbf{\ac{KM}} & \textbf{\ac{WGM}} \\
\midrule
C1 & 9.6 & 13.0 & 11.0 \\
C2 & 23.2 & 24.0 & 21.0 \\
C3 & 30.9 & 32.0 & 28.0 \\
C4 & 8.9 & 10.0 & 8.0 \\
\bottomrule
\end{tabular}
\end{table}

All values exceed $\tau = 7.5$, confirming that no target is reachable in simulation. Interestingly, \ac{WGM} yields the smallest $\tau_{\min}$ for most targets, yet both physics-based models, \ac{KM} and \ac{WGM}, reach the same real-world success of 56.25\%, while \ac{lerp} attains comparably small $\tau_{\min}$ values but fails on every target. This indicates that transfer depends on dynamics accuracy, not just spectrum coverage.

\begin{table}[t]
\centering
\caption{Sim-to-Real Transfer Results. $\bar d_R \pm \sigma$: Mean final distance to target with the standard deviation, in the Avg column, $\sigma$ is the mean of the per-target standard deviations. $s_R$: Steps taken in reality, Succ: Success in each run in \%.}
\label{tab:results_detailed}
\begin{tabular}{l l c c|c}
\toprule
\textbf{Model} & \textbf{Metric} & C1 & C4 & \textbf{Avg}\\
\midrule
\multirow{4}{*}{M1} 
  & $\bar d_R \pm \sigma$ & 12.98 $\pm$ 2.28 & 42.32 $\pm$ 2.62 & \textbf{17.10$\pm$1.79}\\
  & $s_R$     & 5   & 5   & \textbf{4}\\
  & Succ (\%) & 0   & 0   & \textbf{43.75}\\
\midrule
\multirow{4}{*}{M2} 
  & $\bar d_R \pm \sigma$ & 108.62 $\pm$ 1.86 & 46.74 $\pm$ 3.39 & \textbf{54.39$\pm$2.13}\\
  & $s_R$     & 5   & 5   & \textbf{5}\\
  & Succ (\%) & 0   & 0   & \textbf{0}\\
\midrule
\multirow{4}{*}{M3} 
  & $\bar d_R \pm \sigma$ & 94.29 $\pm$ 1.83 & 38.20 $\pm$ 1.98 & \textbf{47.90$\pm$2.66}\\
  & $s_R$    & 5   & 5   & \textbf{5}\\
  & Succ (\%) & 0   & 0   & \textbf{0}\\
\midrule
\multirow{4}{*}{M4} 
  & $\bar d_R \pm \sigma$ & 11.97 $\pm$ 1.18 & 26.99 $\pm$ 1.55 & \textbf{12.72$\pm$1.06}\\
  & $s_R$     & 5   & 5   & \textbf{3.55}\\
  & Succ (\%) & 25   & 0   & \textbf{56.25}\\
\midrule
\multirow{4}{*}{M5} 
  & $\bar d_R \pm \sigma$ & 7.14$\pm$2.15 & 20.16$\pm$1.60 & \textbf{12.74$\pm$2.58}\\
  & $s_R$     & 4.5 & 5   & \textbf{3.93}\\
  & Succ (\%) & 100  & 0   & \textbf{56.25}\\
\bottomrule
\end{tabular}
\end{table}

\section{CONCLUSION AND FUTURE WORK}\label{sec:conclusion}

Our study demonstrates that \ac{mdp} formulation is a crucial factor for sim-to-real transfer in color mixing, complementing dynamics-focused approaches. The key findings are:  
\begin{itemize}
    \item \textbf{Target state inclusion} is essential for meaningful transfer.  
    \item \textbf{Physically motivated dynamics models} (\ac{KM},\ac{WGM}) enable robust real-world performance under precision constraints.  
    \item \textbf{Simple, distance-based rewards} and \textbf{relative state representations} stabilize learning and improve generalization.  
    \item \textbf{Episode parameters} interact with dynamics fidelity: strict settings fail for low-fidelity models but succeed with high-fidelity models.  
\end{itemize}

\textbf{Limitations.} Analysis is limited to a single task domain. Color dominance can make the closest RGB start color suboptimal, and composite scoring weights need further validation. Target colors in hardware lie outside the simulation range, preventing direct sim-to-real evaluation.

\textbf{Future Work.} Future work includes calibrating dynamics models to reduce simulation bias. Studies should focus on reachable targets or alternative gap metrics that do not rely on exact target attainment. Additionally, curriculum learning, that is, gradually increasing task complexity during training, may improve sample efficiency and robustness in transfer.

\bibliographystyle{ieeetr}
\bibliography{Paper_MDP_Design}

@book{sutton_reinforcement_2018,
	address = {Cambridge, Massachusetts},
	edition = {Second edition},
	series = {Adaptive computation and machine learning series},
	title = {Reinforcement learning: an introduction},
	isbn = {978-0-262-03924-6},
	publisher = {The MIT Press},
	author = {Sutton, Richard S. and Barto, Andrew G.},
	year = {2018},
}

@article{sochorova_practical_2021,
	title = {Practical pigment mixing for digital painting},
	volume = {40},
	doi = {10.1145/3478513.3480549},
	number = {6},
	journal = {ACM Transactions on Graphics},
  author = {Sochorov\'{a}, {\v{S}}\'{a}rka and Jamri\v{s}ka, Ond\v{r}ej},
	year = {2021},
	pages = {1--11},
}

@article{hofer_sim2real_2021,
	title = {{Sim2Real} in {Robotics} and {Automation}: {Applications} and {Challenges}},
	volume = {18},
	doi = {10.1109/TASE.2021.3064065},
	journal = {IEEE Transactions on Automation Science and Engineering},
	author = {Höfer, Sebastian and Bekris, Kostas and Handa, Ankur and Higuera, Juan and Mozifian, Melissa and Golemo, Florian and Atkeson, Chris and Fox, Dieter and Goldberg, Ken and Leonard, John and Liu, C. and Peters, Jan and Song, Shuran and Welinder, Peter and White, Martha},
	year = {2021},
	pages = {398--400},
}

@inproceedings{krau_exploring_variance,
	author = {Krau, Tatjana and Schäfer, Georg and Damm, Tobias and Heieck, Frieder},
	year = {2025},
	month = {05},
    booktitle = {2025 10th International Conference on Machine Learning Technologies (ICMLT)},
	pages = {150-157},
	title = {Exploring Variance in Reinforcement Learning Environment to Bridge the Sim-To-Real Gap: A Case Study in Color Mixing},
	doi = {10.1109/ICMLT65785.2025.11193289}
}

@misc{schafer_crucial_2025,
	title = {The {Crucial} {Role} of {Problem} {Formulation} in {Real}-{World} {Reinforcement} {Learning}},
	url = {http://arxiv.org/abs/2503.20442},
	doi = {10.48550/arXiv.2503.20442},
	publisher = {arXiv},
	author = {Schäfer, Georg and Krau, Tatjana and Rehrl, Jakob and Huber, Stefan and Hirlaender, Simon},
	year = {2025},
	note = {Accepted at ICPS 2025},
}

@misc{schulman_proximal_2017,
	title = {Proximal {Policy} {Optimization} {Algorithms}},
	url = {http://arxiv.org/abs/1707.06347},
	doi = {10.48550/arXiv.1707.06347},
	publisher = {arXiv},
	author = {Schulman, John and Wolski, Filip and Dhariwal, Prafulla and Radford, Alec and Klimov, Oleg},
	year = {2017},
}

@article{raffin_stable-baselines3_2021,
	title = {Stable-Baselines3: Reliable Reinforcement Learning Implementations},
	volume = {22},
	journal = {Journal of Machine Learning Research},
	author = {Raffin, Antonin and Hill, Ashley and Gleave, Adam and Kanervisto, Anssi and Ernestus, Maximilian and Dormann, Noah},
	year = {2021},
	pages = {1--8},
}

@article{simonet_WGM,
	author = {Simonot, Lionel and Hébert, Mathieu},
	year = {2014},
	pages = {58--66},
	title = {Between additive and subtractive color mixings: intermediate mixing models},
	volume = {31},
	journal = {Journal of the Optical Society of America A},
	doi = {10.1364/JOSAA.31.000058}
}

@inproceedings{ghosh_why_2021,
	title = {Why {Generalization} in {RL} is {Difficult}: {Epistemic} {POMDPs} and {Implicit} {Partial} {Observability}},
	booktitle = {Advances in Neural Information Processing Systems (NeurIPS)},
	author = {Ghosh, Dibya and Rahme, Jad and Kumar, Aviral and Zhang, Amy and Adams, Ryan P. and Levine, Sergey},
	year = {2021},
	pages = {25502--25515},
}

@inproceedings{schaul_universal_2015,
	title = {Universal Value Function Approximators},
	booktitle = {Proceedings of the 32nd International Conference on Machine Learning (ICML)},
	author = {Schaul, Tom and Horgan, Dan and Gregor, Karol and Silver, David},
	year = {2015},
	pages = {1312--1320},
	publisher = {PMLR}
}

@inproceedings{kaelbling_learning_1993,
	title = {Learning to Achieve Goals},
	booktitle = {Proceedings of the 13th International Joint Conference on Artificial Intelligence (IJCAI)},
	author = {Kaelbling, Leslie Pack},
	year = {1993},
	pages = {1094--1098}
}

@article{recht_tour_2019,
	title = {A Tour of Reinforcement Learning: The View from Continuous Control},
	journal = {Annual Review of Control, Robotics, and Autonomous Systems},
	volume = {2},
	pages = {253--279},
	author = {Recht, Benjamin},
	year = {2019},
	doi = {10.1146/annurev-control-053018-023825}
}

@article{gros_datadriven_2020,
	title = {Data-Driven Economic {NMPC} Using Reinforcement Learning},
	journal = {IEEE Transactions on Automatic Control},
	volume = {65},
	number = {2},
	pages = {636--648},
	author = {Gros, Sébastien and Zanon, Mario},
	year = {2020},
	doi = {10.1109/TAC.2019.2913768}
}

@misc{huang_adversarial_2017,
	title = {Adversarial {Attacks} on {Neural} {Network} {Policies}},
	url = {http://arxiv.org/abs/1702.02284},
	doi = {10.48550/arXiv.1702.02284},
	urldate = {2026-01-19},
	publisher = {arXiv},
	author = {Huang, Sandy and Papernot, Nicolas and Goodfellow, Ian and Duan, Yan and Abbeel, Pieter},
	month = feb,
	year = {2017},
	note = {arXiv:1702.02284 [cs]},
}

\end{document}